\documentclass{article}

\usepackage[final]{neurips_2019}

\usepackage[utf8]{inputenc}
\usepackage[T1]{fontenc}
\usepackage{hyperref}
\usepackage{url}
\usepackage{booktabs}
\usepackage{amsfonts}
\usepackage{nicefrac}
\usepackage{microtype}
\usepackage{graphicx}
\usepackage{xcolor}
\usepackage{lipsum}

\title{
  Improving Pediatric Emergency \\ Department 
  Triage with Modality Dropout in \\ 
  Late Fusion  Multimodal EHR Models\\
}

\author{
  Tyler Yang \\
  Department of Computer Science \\
  Stanford University \\
  \texttt{tyyang@stanford.edu} \\
  \And
  Romal Mitr \\
  Department of Bioengineering \\
  Stanford University \\
  \texttt{romalm@stanford.edu} \\
}

\begin{document}

\maketitle

\begin{abstract}
Emergency department triage relies heavily on both quantitative vital signs and qualitative clinical notes, yet multimodal machine learning models predicting triage acuity often suffer from modality collapse by over-relying on structured tabular data. This limitation severely hinders demographic generalizability, particularly for pediatric patients where developmental variations in vital signs make unstructured clinical narratives uniquely crucial. To address this gap, we propose a late-fusion multimodal architecture that processes tabular vitals via XGBoost and unstructured clinical text via Bio\_ClinicalBERT, combined through a Logistic Regression meta-classifier to predict the 5-level Emergency Severity Index. To explicitly target the external validity problem, we train our model exclusively on adult encounters from the MIMIC-IV and NHAMCS datasets and evaluate its zero-shot generalization on a traditionally overlooked pediatric cohort. Furthermore, we employ symmetric modality dropout during training to prevent the ensemble from overfitting to adult-specific clinical correlations. Our results demonstrate that the multimodal framework significantly outperforms single-modality baselines. Most notably, applying a 30–40\% symmetric modality dropout rate yielded steep performance improvements in the unseen pediatric cohort, elevating the Quadratic Weighted Kappa to 0.351. These findings highlight modality dropout as a critical regularization technique for mitigating modality collapse and enhancing cross-demographic generalization in clinical AI.
\end{abstract}

\section{Introduction}

Emergency department (ED) triage is critical to patient care, yet current practices are highly variable.
Triage decisions typically rely on a combination of quantitative vital signs and qualitative nurse-
reported notes \citep{yancey2023}.

While machine learning models have been developed to predict Emergency Severity Index (ESI) levels
from multimodal EHR data, many suffer from modality collapse, over-weighting structured vitals
while underutilizing informative clinical text \citep{wang2023}. This limitation is especially problematic in pediatrics, where normal vital ranges vary with age and development, and the most important information is
often conveyed through the nurse descriptions.

To address this gap, we propose a model to improve pediatric triage that jointly leverages structured and unstructured data, while using modality dropout to ensure that neither modality dominates. Our input to the algorithm will be patient vitals and the clinical note at the time of triage. We then use a late-fusion multimodal framework that consists of a gradient boosted decision tree (XGBoost) to process the patient vitals and a transformer (Bio\_ClinicalBERT) to process the clinical note. These two components are brought together through a Logistic Regression model that serves as the final ensemble layer in order to output the predicted triage level (from 1-5).

\section{Related Work}

Recent applications of machine learning in healthcare have increasingly leveraged multimodal medical data, though often with significant limitations in scope or demographic generalizability. For instance, \citet{wang2024} systematically evaluated early, joint, and late fusion strategies to predict hospital readmissions and length of stay. However, their model was trained and tested exclusively on adult patients, limiting its external validity. Similarly, \citet{nover2025} combined XGBoost and Bio-Clinical BERT to synthesize tabular vitals and clinical notes, but framed hospital admission as a simple binary classification rather than a nuanced multi-class problem. \citet{lu2025} created a late-fusion transformer model that focused on pediatric populations, but was predicting pediatric cardiac arrest rather than triage. 

Within the specific domain of multi-class triage, \citet{liu2025} successfully utilized a prompt-tuning system to predict exact triage levels. Researchers have also sought to improve multimodal robustness against missing data. \citet{gu2025} championed an asymmetric modality dropout approach that forces models to learn distributed representations without over-relying on a single modality.

Our project bridges these gaps by synthesizing state-of-the-art late-fusion multimodal architectures with modality dropout to predict multi-class triage levels. Unlike previous studies that suffer from narrow demographic generalizability, we explicitly target the external validity problem by intentionally training on purely adult data and evaluating it on the traditionally overlooked pediatric cohort.

\section{Dataset and Features}

After an application and credentialing process, we obtained access to data through the MIMIC-IV and MIMIC-IV-ED datasets \citep{johnson2023}. We also pulled data from the 2020, 2021, and 2022 NHAMCS datasets \citep{nhamcs2020, nhamcs2021, nhamcs2022}. 

Data preprocessing for MIMIC involved extraction of relevant features by merging databases via $\texttt{subject\_id}$ and interpretation of non-numerical pain ratings. Data preprocessing for NHAMCS involved selection of features to match MIMIC and interpretation of non-numerical ages. We added columns that flag patient gender and whether the patient was able to provide a pain rating. After combining the datasets, we filtered out all rows missing text data and performed outlier removal by thresholding the values so that they are within a physiologically realistic range. Since XGBoost can handle missing values, we did not filter based on missing tabular values. We then split the data into an adult and pediatric cohort. The original MIMIC and NHAMCS datasets contained 425,011 and 47,092 encounters, respectively (472,103 total). After preprocessing, 415,536 MIMIC and 30,282 NHAMCS encounters remained (445,818 total). Of these, 440,237 were adults and 5,581 were pediatric patients.

The text modality feature consisted solely of the $\texttt{chief\_complaint}$ field. Tabular features included $\texttt{gender}$, $\texttt{age\_at\_visit}$, $\texttt{temperature}$, $\texttt{heartrate}$, $\texttt{resp\_rate}$, $\texttt{pain\_score}$, $\texttt{o2\_sat}$, $\texttt{systolic\_bp}$, $\texttt{diastolic\_bp}$, and $\texttt{unable}$ (indicating whether the patient was unable to provide a pain rating). The prediction label was $\texttt{acuity}$, which represents the triage level.

\section{Methods}

We divided the modeling phase into three structured experiments to compare performance: (1) single-modality baselines (benchmarks for tabular and clinical models separately), (2) late-fusion multimodal model, and (3) modality-dropout experiments. We used stratified train–test splitting so that each triage level was proportionally represented in both sets, preventing class imbalance.

\textbf{Baseline single modality tabular model:} We trained an XGBoost classifier to predict triage level (\texttt{acuity}). XGBoost is a gradient-boosted decision tree ensemble method that iteratively minimizes a loss function by combining "weak learners" (shallow decision trees) iteratively. It is an additive algorithm that creates a new tree ($f_t$) one iteration at a time:  $\hat{y}_i^{(t)} = \hat{y}_i^{(t-1)} + f_t(x_i)$, where $\hat{y}_i^{(t)}$ is the predicted y-value at iteration $t$. The tree is formed in order to minimize the following objective function: 
\[
\sum_{i=1}^{n} l\left(y_i, \hat{y}_i^{(t-1)} + f_t(x_i)\right) + \omega(f_t), 
\] 
where $\omega(f_t)$ is the regularization term to balance tree complexity and prevent overfitting \citep{xg2025}. We first trained a standard mutliclass XGBoost that viewed acuity levels as distinct buckets. However, since this does not factor in the context of severity, our second approach was to convert this to an ordinal regression task and use XGBoost to predict a real number in the continuous range from 1 to 5. \citep{kahl2025}. 

\textbf{Baseline single modality text model:} First, we trained a traditional TF-IDF–based classifier. This method works by assigning a weight to each word $w$ based on term frequency (tf) and inverse document frequency (idf) vectors. The tf is found by looking at the frequency of a word in a specific clinical note $d$. The idf prioritizes words that frequently occur in a certain clinical note but are not common otherwise across the dataset D: $idf = \ln \frac{\texttt{number of notes in } D}{\texttt{number of notes containing } w}$. Therefore, the overall score can be calculated by doing: $tf(w, d) \times idf(w, D).$ Based on these weights, it forms a vector that can then be inputted to a supervised classifier \citep{silge2025}. 

Second, we fine-tuned BioClinicalBERT, a transformer model pretrained on medical corpora and clinical notes, to predict triage level from unstructured text. Unlike TF-IDF which looks at each word in isolation, BERT uses self-attention, which means that it considers each word in relation to other words in the sequence \citep{sounack2025}. Self-attention is calculated as a scaled dot product based on query (Q), key (K), and value (V) matrices: 
$$\texttt{Attention}(Q, K, V) = \texttt{softmax} \left( \frac{QK^\top}{\sqrt{d_k}} \right) V,$$
where $d_k$ is the dimension of the keys and helps scale the attention so that the gradients do not vanish \citet{wagner2020}. BERT tokenizes the input and creates a final CLS token based on the entire clinical note. This is used as the input into a sequence classification head for the supervised fine-tuning layer, which then is fed through a softmax function that provides the probability of a certain triage score. 

\textbf{Multimodal:}  The multimodal framework integrates structured tabular features and clinical text using late-fusion. It combines modality-specific predictions by concatenating the projected triage probabilities from the softmax function of our baseline XGBoost and BioClinicalBERT models. Let $p_i ^{tab}$ be the probability vector from the tabular data and $p_i ^{text}$ be the probability vector from the text, then $a_i = [p_i ^{tab}, p_i ^{text}]$. The model then uses a Logistic Regression model to output a final triage prediction for class $c$ based on the concatenated probability vector using the following formula:
\[
P(y_i = c | \mathbf{a}_i) = \frac{e^{w_c^\top \mathbf{a}_i + b_c}}{\sum_{j=1}^{5} e^{w_j^\top \mathbf{a}_i + b_j}}, \texttt{where $w_c$ is the learned weights for class $c$}.
\] 

\textbf{Modality Dropout:} To ensure robustness and interpretability, we evaluated model behavior under varying levels of modality dropout. Our primary analysis focused on symmetric dropout configurations of 10\%, 20\%, 30\%, and 40\%, where both modalities were dropped with equal probability during training. In addition, we explored asymmetric dropout configurations, testing all modality pairs from 0.1 to 0.8 in increments of 0.1.

\section{Experiments}

In our late-fusion multimodal architecture, the XGBoost and Bio\_ClinicalBERT base models utilized the same hyperparameters as their respective single-modality baselines. For the tabular XGBoost model, we configured $\texttt{n\_estimators}=500$ and $\texttt{learning\_rate}=0.05$, a well-documented combination in gradient boosting literature that balances robust learning with sufficient ensemble capacity \citep{chen2016}. To further regularize this process, we implemented $\texttt{early\_stopping\_rounds}=25$. We explicitly set $\texttt{eval\_metric}$='mlogloss' to evaluate this stopping criterion, as it is appropriate for optimizing multi-class probability distributions. Transitioning to the text modality, the Bio\_ClinicalBERT model was constrained to $\texttt{max\_length}=32$; during text processing, we found that the maximum token length of our chief complaints was 16, making 32 more than enough to capture the full context while drastically saving memory and compute time. We paired this with a $\texttt{learning\_rate}$ of 2e-5 and a $\texttt{per\_device\_train\_batch\_size}$ of 32, which is widely recognized as a classic, highly stable hyperparameter combination for fine-tuning BERT architectures \citep{devlin2018}. For the TF-IDF model, we selected an n-gram range of \texttt{(1,3)} and \texttt{C = 0.1} because this configuration achieved the highest macro-F1 score on the validation set. Finally, for our Logistic Regression meta-classifier, we observed that the default solver failed to converge within 100 iterations on the stacked probability arrays, prompting us to increase $\texttt{max\_iter}$ to 1000 to ensure stable gradient descent. We maintained the default L2 regularization ($\texttt{penalty}$='l2', $C$=1.0) to prevent the model from over-weighting any single predicted probability from the base models.

\begin{figure}[hbt!]
    \centering
    \includegraphics[width=0.9\linewidth]{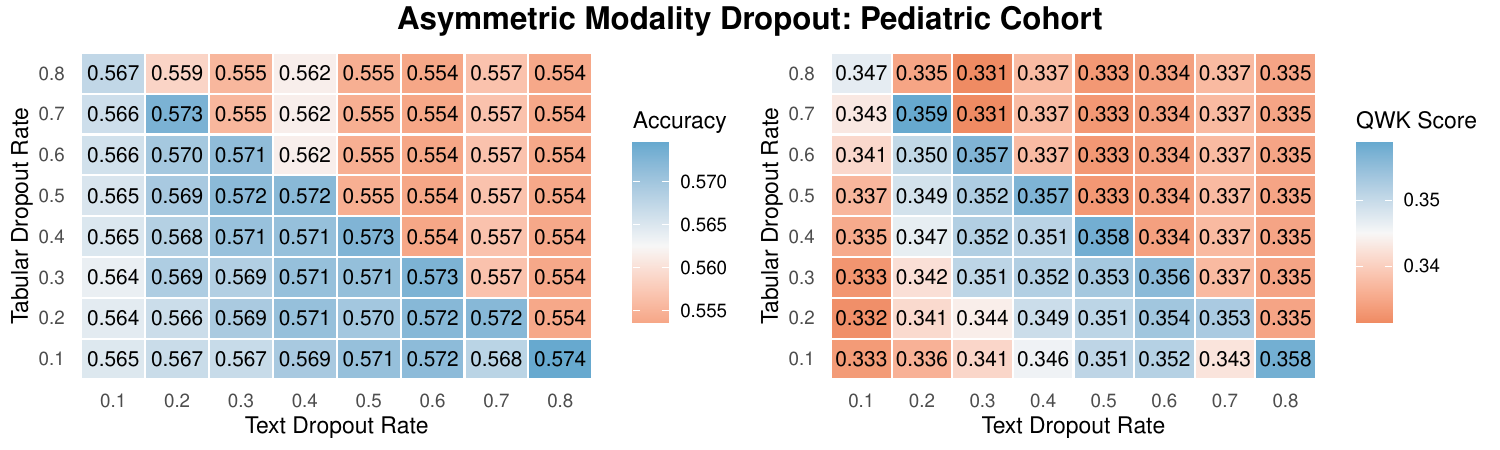}
    \caption{\textbf{Heatmap of multimodal model performance across asymmetric modality dropout combinations.} The heatmaps display the overall Accuracy (left panel) and Quadratic Weighted Kappa (right panel) achieved by the late fusion multimodal model when subjected to independent dropout rates for the tabular and text modalities. Dropout probabilities for tabular data and clinical text each range from 0.1 to 0.8 in increments of 0.1. Cell color intensity corresponds to predictive performance.}
    \label{fig1}
\end{figure}

During this final ensemble phase, we also experimented with various symmetric and asymmetric modality dropout rates to evaluate their impact on generalization. In exploratory analyses, we evaluated the impact of asymmetric modality dropout, wherein independent dropout probabilities ranging from 0.1 to 0.8 were assigned to the text and tabular data streams. As illustrated in Figure \ref{fig1}, while certain specific asymmetric combinations yielded marginally higher Kappa scores on the validation set, there were no consistent patterns. Thus, we elected to utilize symmetric modality dropout (a single, uniform rate across all modalities) for our primary analyses and final model selection. This streamlined approach reduces hyperparameter tuning overhead and guards against overfitting to specific modalities, ensuring a more stable and generalizable model.

We utilized a 60-20-20 train-validation-test split, which helped us tune the base models and train the Level 2 meta-classifier. This stacking approach enabled the ensemble to optimally weight the text and tabular modalities based on their generalization performance.

\section{Results / Discussion}

Model performance was primarily evaluated using the Quadratic Weighted Kappa (QWK). Unlike standard accuracy, QWK is well-suited for ordinal clinical scales like emergency triage because it applies a penalty proportional to the square of the distance between predicted and actual categories \citep{hermann1996}. By penalizing catastrophic errors while correcting for chance agreement, QWK serves as the most clinically relevant metric for modern AI triage evaluation \citep{nedos2026, wong2026}. The equation for QWK is

$$\kappa_w = 1 - \frac{\sum_{i=1}^{k} \sum_{j=1}^{k} w_{ij} O_{ij}}{\sum_{i=1}^{k} \sum_{j=1}^{k} w_{ij} E_{ij}},$$

where $O_{ij}$ is the observed proportion of cases where rater A assigned label $i$ and rater B assigned label $j$, $E_{ij}$ is the expected proportion of cases by chance, and $w_{ij}= \frac{(i - j)^2}{(k - 1)^2}$ is the weight assigned to each cell based on the distance between categories. Secondary metrics include accuracy, balanced accuracy, and Macro F1 Score. 

\begin{figure}[hbt!]
    \centering
    \includegraphics[width=0.9\linewidth]{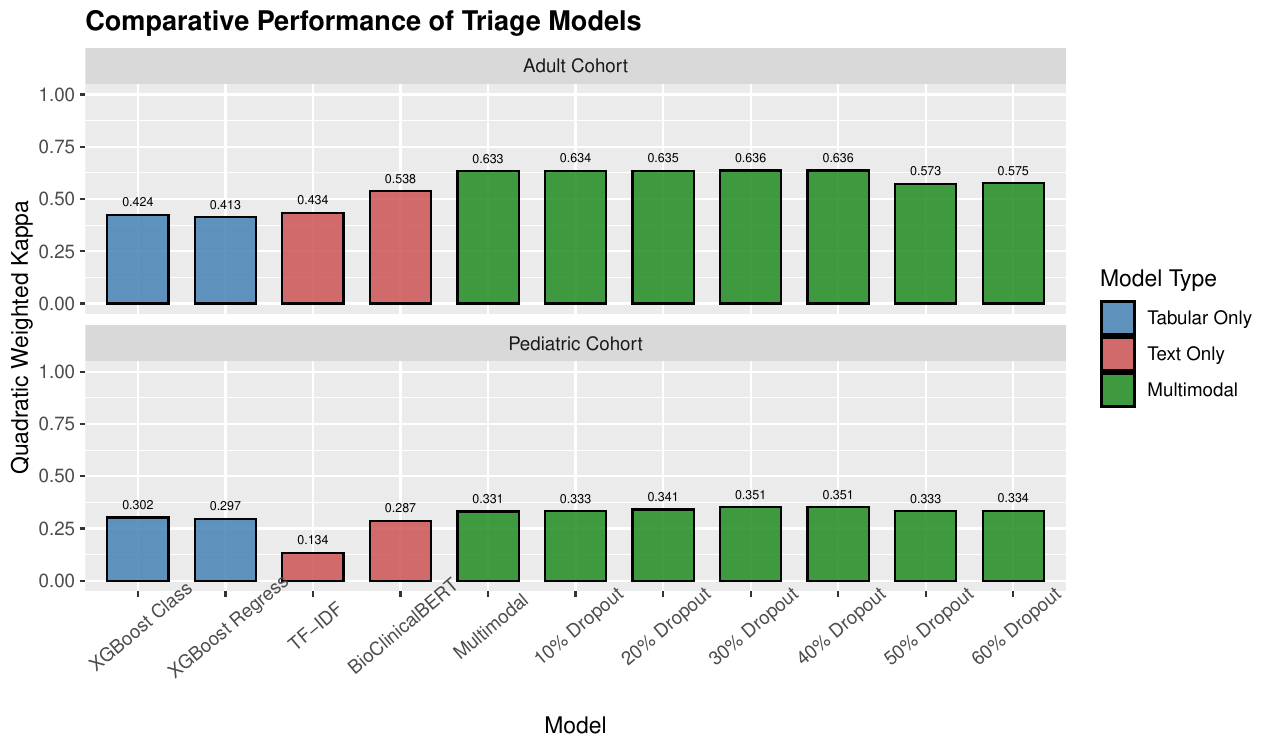}
    \caption{\textbf{Comparative model performance on adult \& pediatric cohorts}. Performance (QWK) is grouped by input modality: Tabular Only (blue), Text Only (red), and Multimodal (green). The multimodal group includes the baseline and variants using modality dropout rates from 10\% to 60\%.}
    \label{fig2}
    
\end{figure}

Consistent with our hypotheses, the multimodal approach outperformed the single-modality baselines for both adult and pediatric cases. In the adult cohort, the baseline multimodal model achieved a QWK of 0.633, representing a distinct improvement over the best single-modality baselines (0.424 for tabular, 0.538 for text). Similarly, for pediatric cohorts, the no-dropout multimodal model achieved a QWK of 0.331, which was an improvement compared to 0.302 and 0.287 for tabular and text baselines. \\

Furthermore, the application of modality dropout yielded additional performance gains. Peak performance was observed at a 30\% to 40\% dropout rate for both cohorts, reaching a Kappa of 0.636 in the adult group and 0.351 in the pediatric group (Figure \ref{fig2}). Thus, the 40\% dropout configuration was selected for subsequent analyses, and its confusion matrices are shown in Tables 1 and 2. As shown in Figure \ref{fig3}, the dropout-based improvements were greater in the pediatric cohort. While performance in the adult cohort remained relatively flat and stable across all four metrics up to the 40\% threshold, the pediatric cohort demonstrated a distinct upward trajectory. Notably, both cohorts exhibited a steep drop-off in performance when the dropout rate reached 50\%. This suggests that while moderate dropout serves as an effective regularization technique, excessive dropout starves the model of the information necessary to learn effective representations. 

\begin{figure}[hbt!]
    \centering
    \includegraphics[width=0.8\linewidth]{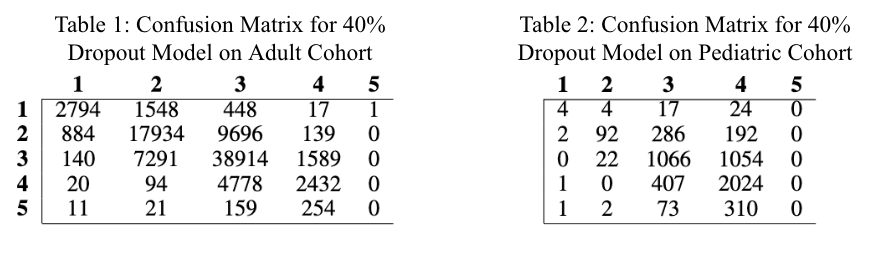}
    \label{fig:placeholder}
\end{figure}

\begin{figure}[hbt!]
    \centering
    \includegraphics[width=0.7\linewidth]{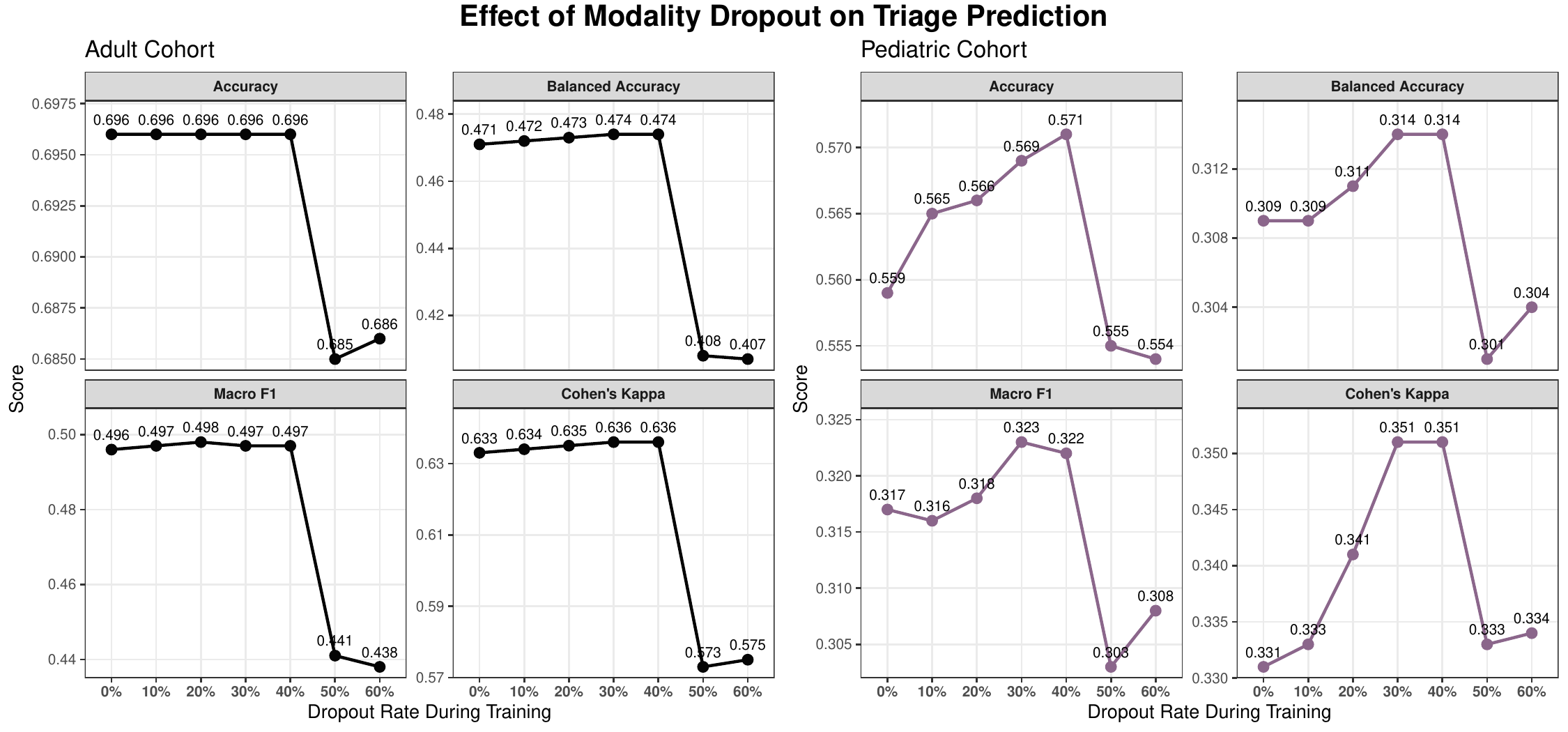}
    \caption{\textbf{Impact of modality dropout on model performance metrics.} The line graphs display Accuracy, Balanced Accuracy, Macro F1, and Cohen's Kappa across training dropout rates from 0\% to 60\%. Results are plotted in black for the Adult Cohort and in purple for the Pediatric Cohort.}
    \label{fig3}
\end{figure}

\begin{figure}
    \centering
    \includegraphics[width=0.8\linewidth]{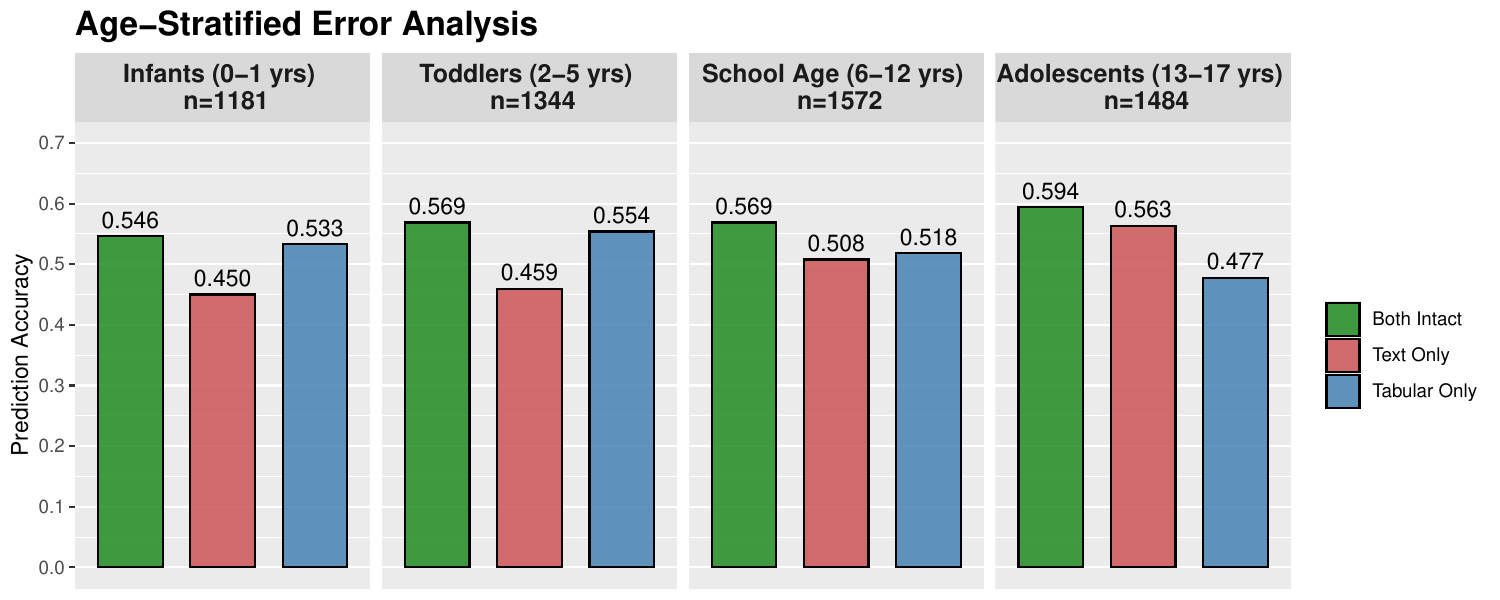}
    \caption{\textbf{Age-stratified error analysis within the pediatric cohort.} Prediction accuracy of the meta-classifier (trained with 40\% modality dropout) across four clinical cohorts: Infants (0–1 yrs), Toddlers/Preschool (2–5 yrs), School Age (6–12 yrs), and Adolescents (13–17 yrs). "Both Intact" represents the full multimodal input, while "No Tabular" and "No Text" reflect model accuracy when specific feature sets are zero-masked to simulate missing clinical vitals or narrative notes, respectively.}
    \label{fig4}
\end{figure}

It is worth noting the disparity in overall performance between the two populations. While the models demonstrated predictive utility in the adult cohort, the absolute values in the pediatric cohort remained comparatively low across all architectures, improving slightly with modality dropout (Figures \ref{fig2} and \ref{fig3}). Age-stratified error analysis shown in Figure \ref{fig4} reveals that this performance penalty is not uniformly distributed across pediatric age groups or input modalities. While the intact multimodal model consistently achieves the highest accuracy across all age brackets, the relative contribution of each modality shifts significantly as patient age increases. In the youngest populations (Infants and Toddlers), the adult-trained model relies heavily on tabular data, with the tabular-only baseline vastly outperforming the text-only baseline. Conversely, in the Adolescent group (13–17 yrs), this trend completely inverses: text-only prediction accuracy improves dramatically to 0.563, while tabular-only performance drops to its lowest point across all cohorts (0.477). Ultimately, the intact multimodal performance improves incrementally with patient age, peaking at 0.594 in adolescents, which likely reflects their closer approximation to the adult clinical profiles upon which the model was trained.

\section{Conclusions}
We evaluated the ED triage performance and generalizability of unimodal and multimodal frameworks under the realistic clinical constraint of training exclusively on adult data but assessing on pediatric data. Our findings demonstrate that late-fusion multimodal architectures outperform single-modality baselines, likely because synthesizing modalities provides a more holistic representation. However, performance significantly degraded when tested on pediatric patients.

To address this gap, we employed modality dropout during the adult-only training phase. While moderate symmetric modality dropout (up to 40\%) yielded marginal stability gains in the adult test set, it drove much steeper performance improvement in unseen pediatric cohorts. This suggests that modality dropout is a critical regularization technique for pediatric generalization, preventing the model from over-fitting to adult-specific correlations between vital signs and clinical language. By forcing it to learn robust, independent representations from each modality, the model can generalize well to varying age groups. Exploratory analyses showed no consistent benefit from asymmetric dropout configurations, suggesting that a streamlined, symmetric approach provides the most robust regularization without adding unnecessary tuning complexity. Future work will explore early-fusion architectures as well as identifying the specific clinical variables driving this degradation and developing targeted debiasing techniques.

\section{Acknowledgments}

We gratefully acknowledge the Stanford Department of Computer Science, as well as PhysioNet \citep{goldberger2000} and Centers of Disease Control and Prevention \citep{nhamcs2020, nhamcs2021, nhamcs2022} for providing access to several of the clinical datasets used in this study. 

\section{Code availability statement}

We provide the \href{https://github.com/tyyang05/peds_triage}{source code} for all of our models for reproducibility on Github.

\bibliographystyle{acl_natbib}

\bibliography{references}
\clearpage
\appendix
\section{Appendix}

\subsection{Appendix A: Extended Performance Metrics}

\begin{table}[h!]
\centering
\begin{tabular}{l r r r r r r}
\hline
\multicolumn{7}{c}{\textbf{Performances on Adult Cohort}} \\
\hline
\textbf{Model} & \textbf{Training Error} & \textbf{Test Error} & \textbf{QWK} & \textbf{Accuracy} & \textbf{Balanced Acc} & \textbf{Macro F1} \\
\hline
XGBoost Class & 0.888 & 0.912 & 0.424 & 0.600 & 0.360 & 0.370 \\
XGBoost Regress & 0.612 & 0.621 & 0.413 & 0.600 & 0.340 & 0.350 \\
TF-IDF & 1.74 & 1.74 & 0.434 & 0.498 & 0.491 & 0.364 \\
BioClinicalBERT & 0.762 & 0.778 & 0.538 & 0.654 & 0.410 & 0.433 \\
Multimodal & 0.721 & 0.724 & 0.633 & 0.696 & 0.471 & 0.496 \\
10\% Dropout & 0.721 & 0.724 & 0.634 & 0.696 & 0.472 & 0.497 \\
20\% Dropout & 0.722 & 0.725 & 0.635 & 0.696 & 0.473 & 0.498 \\
30\% Dropout & 0.724 & 0.727 & 0.636 & 0.696 & 0.474 & 0.497 \\
40\% Dropout & 0.726 & 0.73 & 0.636 & 0.696 & 0.474 & 0.497 \\
50\% Dropout & 0.782 & 0.785 & 0.573 & 0.685 & 0.408 & 0.441 \\
60\% Dropout & 0.782 & 0.785 & 0.575 & 0.686 & 0.407 & 0.438 \\
\hline
\end{tabular}
\end{table}
\begin{center}
Table A1: \textbf{Extended Performance Metrics on the Adult Cohort.}
\end{center}

Table A1 details the comprehensive evaluation metrics for all models tested on the Adult Cohort, including Training and Test Errors, Quadratic Weighted Kappa (QWK), Accuracy, Balanced Accuracy, and Macro F1 scores. The results demonstrate that the advanced architectures, particularly the multimodal models, consistently outperform traditional baselines and the unimodal BioClinicalBERT across primary classification metrics. Furthermore, the table includes an ablation study evaluating the impact of varying dropout rates on the Multimodal model. The data indicates that the model maintains robust and optimal performance with dropout rates between 10\% and 40\%—achieving peak QWK and Macro F1 scores within this range—before experiencing a noticeable degradation in performance at the 50\% and 60\% dropout thresholds.

\begin{table}[h!]
\centering
\begin{tabular}{l r r r r}
\hline
\multicolumn{5}{c}{\textbf{Performances on Pediatric Cohort}} \\
\hline
\textbf{Model} & \textbf{QWK} & \textbf{Accuracy} & \textbf{Balanced Acc} & \textbf{Macro F1} \\
\hline
XGBoost Class & 0.302 & 0.560 & 0.340 & 0.380 \\
XGBoost Regress & 0.297 & 0.510 & 0.260 & 0.250 \\
TF-IDF & 0.134 & 0.325 & 0.272 & 0.226 \\
BioClinicalBERT & 0.287 & 0.481 & 0.285 & 0.284 \\
Multimodal & 0.331 & 0.559 & 0.309 & 0.317 \\
10\% Dropout & 0.333 & 0.565 & 0.309 & 0.316 \\
20\% Dropout & 0.341 & 0.566 & 0.311 & 0.318 \\
30\% Dropout & 0.351 & 0.569 & 0.314 & 0.323 \\
40\% Dropout & 0.351 & 0.571 & 0.314 & 0.322 \\
50\% Dropout & 0.333 & 0.555 & 0.301 & 0.303 \\
60\% Dropout & 0.334 & 0.554 & 0.304 & 0.308 \\
\hline
\end{tabular}
\end{table}
\begin{center}
Table A2: \textbf{Extended Performance Metrics on the Pediatric Cohort.}
\end{center}

Similarly, Table A2 presents the evaluation metrics for the models tested on the Pediatric Cohort, detailing their QWK, Accuracy, Balanced Accuracy, and Macro F1 scores. Interestingly, the results here present a slightly different landscape compared to the adult cohort: while the Multimodal architecture with optimized dropout rates (peaking between 30\% and 40\%) achieves the highest QWK (0.351) and Accuracy (0.571), the baseline XGBoost Class model actually retains the highest Balanced Accuracy (0.340) and Macro F1 (0.380) overall. The dropout ablation study, however, mirrors the adult cohort trends, indicating that the Multimodal model's performance begins to degrade once dropout rates reach the 50\% threshold and higher.

\end{document}